# Real-Time Sign Language to text Translation using Deep Learning: A Comparative study of LSTM and 3D CNN

Anvay Anturkar
Vishwakarma Institute of Technology, Pune
Pune- 411037, Maharashtra, India

Anushka Khot
Vishwakarma Institute of Technology, Pune
Pune- 411037, Maharashtra, India

Ayush Andure
Vishwakarma Institute of Technology, Pune
Pune- 411037, Maharashtra, India

Aniruddha Ghosh
Vishwakarma Institute of Technology, Pune
Pune- 411037, Maharashtra, India

Anvit Magadum
Vishwakarma Institute of Technology, Pune
Pune- 411037, Maharashtra, India

Anvay Bahadur
Vishwakarma Institute of Technology, Pune
Pune- 411037, Maharashtra, India

Madhumati Pol
Vishwakarma Institute of Technology, Pune
Pune- 411037, Maharashtra, India

## ABSTRACT
This study investigates the performance of 3D Convolutional Neural Networks (3D CNNs) and Long Short-Term Memory (LSTM) networks for real-time American Sign Language (ASL) recognition. Though 3D CNNs are good at spatiotemporal feature extraction from video sequences, LSTMs are optimized for modeling temporal dependencies in sequential data. Both architectures were evaluated on a dataset containing 1,200 ASL signs across 50 classes, comparing their accuracy, computational efficiency, and latency under similar training conditions. Experimental results demonstrate that 3D CNNs achieve 92.4% recognition accuracy but require 3.2× more processing time per frame compared to LSTMs, which maintain 86.7% accuracy with significantly lower resource consumption. The hybrid 3D CNN-LSTM model shows decent performance, which suggests that context-dependent architecture selection is crucial for practical implementation. This project provides professional benchmarks for developing assistive technologies, highlighting trade-offs between recognition precision and real-time operational requirements in edge computing environments.

## 1. INTRODUCTION
Sign language is the primary way that people in the Deaf and Hard of Hearing (DHH) community communicate, yet there's still a big gap because automated translation tech isn't widely used. Recent advancements in deep learning have facilitated notable progress in the development of vision-based systems capable of recognizing and transcribing sign language into text. Two prominent methodologies include Long Short-Term Memory (LSTM) networks, which are proficient in modeling the temporal sequence of hand movements, and 3D Convolutional Neural Networks (3D CNNs), which process both the visual and time-based aspects of sign language videos all at once. LSTMs are good at tracking the flow of gestures over time, while 3D CNNs excel at capturing both spatial configurations and dynamic movements.

This paper presents a comparative analysis of these two approaches for the translation of individual American Sign Language (ASL) gestures into text. The study evaluates their accuracy, computational requirements, training complexity, and real-time applicability. MediaPipe was employed for reliable hand tracking to ensure consistent preprocessing across models. The study findings provide practical guidance for researchers and developers in the field of sign language recognition technology by delineating the respective advantages and limitations of each method. By tackling the challenges of choosing and optimizing these models, this work aims to help create better, more accessible tools for the DHH community and pave the way for future advancements in translating continuous sign language.

## 2. LITERATURE REVIEW
Necati Cihan Camgoz et al. [1] demonstrated the effectiveness of 3D CNNs in capturing spatiotemporal features from raw sign language videos, highlighting the model's ability to process both spatial (hand shape) and temporal (motion) information simultaneously. Their work emphasized the trade-off between computational complexity and accuracy, as 3D CNNs require high resources but excel in recognizing visually distinct gestures.

M. Al-Qurishi et al. [2] provided a comprehensive survey of deep learning techniques, including 3D CNNs and LSTMs, for sign language recognition. Their study compared performance across benchmarks and discussed key challenges such as dataset scarcity and real-time deployment - issues directly relevant to this research.

Yanqiong Zhang and Xianwei Jiang [3] analyzed modern architectures like 3D CNNs, Transformers, and hybrid models for sign language recognition. Their review highlighted advancements in spatiotemporal modeling and attention mechanisms for improving long-sequence recognition, reinforcing the choice of LSTM for efficient temporal modeling in this study.

Ur Rehman et al. [4] proposed a hybrid deep learning framework combining 3D CNNs and LSTMs to exploit spatial and temporal dependencies. Although the hybrid model achieved superior accuracy, it also introduced significant computational overhead. This finding supported the decision to evaluate both architectures independently for real-time suitability.

X. Ouyang et al. [5] introduced a multi-task learning architecture integrating 3D CNNs and LSTMs for action recognition. Their model jointly optimized spatial and temporal learning but exhibited high computational cost, emphasizing the trade-off between accuracy and feasibility - a central consideration of the present work.

Dushyant Kumar Singh [6] demonstrated the effectiveness of 3D CNNs in recognizing dynamic gestures within Indian Sign Language (ISL). The study highlighted the model's strong spatial-temporal learning capabilities alongside high resource dependency, aligning with similar observations made in this comparative analysis between the models effectiveness.

Ma et al. [7] proposed an attention-based 3D CNN model that enhanced focus on salient spatiotemporal features in sign language videos, achieving a 92.3% recognition rate with ~45 ms GPU latency. Their approach validated the importance of attention mechanisms in improving interpretability and real-time performance of the proposed 3D CNN Model.

P. Sinha et al. [8] examined a CNN-LSTM hybrid architecture for real-time sign prediction, noting improved accuracy but substantial computational overhead. Their findings motivated this study's evaluation of standalone 3D CNN and LSTM architectures for efficiency comparison.

The author in [9] presented an attention-enhanced CNN-LSTM framework achieving state-of-the-art results but with increased processing cost. This work inspired further exploration of lightweight attention optimization for real-time applications.

D. D. Meshram et al. [10] reviewed deep learning approaches for Indian Sign Language, highlighting that 3D CNNs dominate spatial-temporal feature extraction, while attention-based LSTMs excel at continuous gesture recognition. Their analysis underscored the need for lightweight, region-specific models - an objective also addressed by this research.

# 3. METHODOLOGY

The primary objective of this project is to develop an efficient and accurate system for translating sign language gestures into text. The methodology is structured into four main stages: data acquisition, preprocessing, model training, and performance evaluation. We have developed a LSTM (Long Short-Term Memory) Model and compared it with 3D - CNN Model architecture.

**Data Acquisition and Preprocessing:**

This work utilizes publicly available datasets, including the Indian Sign Language (ISL) dataset and an American Sign Language (ASL) dataset. The datasets comprise both static hand postures and dynamic gesture sequences corresponding to alphabets and numerals.

**Steps:**

- Images and sequences of frames were extracted from real-time webcam feed or pre-recorded datasets.
- MediaPipe was used to extract 3D hand landmarks (21 points per hand, each with x, y, z coordinates), resulting in 63 features per frame for single-hand tracking.
- These features were normalized and reshaped to prepare them for time-series or spatial analysis, depending on the model.

## 3.1 LSTM-Based Sign Language Recognition:

The Long Short-Term Memory (LSTM) network, a type of Recurrent Neural Network (RNN), is particularly well-suited for sequence prediction problems, especially when there are long-term dependencies across time steps. In the context of sign language recognition, gestures are basically sequential-coming in sequences, a sign is not just a static posture but also use of hand movements over time. LSTM networks are capable of learning and remembering this information, making them highly effective for dynamic gesture recognition problems. The goal of the LSTM model in our system is to interpret a continuous stream of hand gestures captured in real-time from a webcam and convert them into corresponding alphabets or numbers. The model uses sequences of hand landmark coordinates, which are numerical representations of the spatial position of each key point on the hand across time.

### *3.1.1 Model Architecture:*

The model is made of LSTM (Long Short-Term Memory) layers, capturing the entire trajectory of a hand gesture rather than just its position at one moment. These layers are designed to work as a team: the lower layers zero in on subtle details, like slight changes in finger positioning or wrist angles, while the higher layers build on this to understand the broader "movement signature" that defines a specific gesture, such as the fluid motion of signing a letter or number.

To ensure the model doesn't just memorize the training examples and can adapt to new, unseen gestures, a Dropout layer was included right after the LSTMs. During training, this layer randomly deactivates a portion of neurons, forcing the model to learn more flexible patterns. It's like training the network to stay sharp even when some of its tools are temporarily unavailable, which helps it generalize better and perform reliably on fresh data.

After the LSTM layers, fully connected (Dense) layers are employed to transform the temporal features into high-dimensional representations for classification. The final output layer applies a softmax activation function with $C$ units, where $C$ corresponds to the number of gesture classes (e.g., 36 for alphabets A-Z and digits 0-9). This produces a probability distribution over all gesture classes

At the end of the architecture, there is the Output layer, which uses a softmax activation function. This layer is designed with exactly as many units as there are gesture classes to recognize

- for example, 36 units to cover the alphabet (A-Z) and numbers (0-9). The softmax layer processes the Dense layer outputs and generates a probability distribution, giving a confidence score for each possible gesture. This means that for any given hand movement, the model not only identifies the most likely character but also provides a sense of how certain it is about each potential match, making it easier to trust and interpret its predictions.

### *3.1.2 Advantages of LSTM Model:*

- Temporal Awareness:
  Unlike traditional CNNs which only analyze spatial features, LSTM models inherently understand the temporal evolution of gestures.
- Handles Variable-Length Inputs:
  LSTM can process sequences of varying lengths, making it robust to different gesture speeds and durations.
- Real-Time Capability:
  The model's relatively small computational footprint allows it to run in real time on standard consumer hardware without requiring a GPU.
- Noise Tolerance:
  Since the input is based on 3D hand landmarks rather than raw pixel data, the model is less sensitive to background noise and lighting variations, improving robustness in diverse environments
- Scalability:
  The model can be easily extended to learn phrases or full sign language sentences by feeding longer sequences or stacking gesture outputs.

## 3.2 3D Convolutional Neural Network (3D-CNN) Model:

While LSTMs excel at learning temporal dependencies in sequential data, Convolutional Neural Networks (CNNs) particularly 3D CNNs offer a powerful alternative by learning spatiotemporal features directly from raw video input. A 3D CNN applies convolutional filters across both spatial dimensions (height, width) and the temporal dimension (time), making it especially suitable for video classification tasks where both motion and appearance are important.

In this research, 3D CNN architecture was used as a comparative baseline to evaluate how well a spatial-temporal convolutional approach performs against the LSTM model for real-time sign language gesture recognition.

### *3.2.1 Model Architecture:*

The 3D convolutional layers act as the core feature extractors in the model. They process short video clips using volumetric kernels that look at both the spatial layout (what's happening in each frame) and the temporal flow (how things change over time). For instance, a 3×3×3 kernel analyzes a small 3×3 area across three consecutive frames, helping the model understand not just the shape of the hands but also how they move - both of which are crucial for recognizing signs.

Multiple 3D convolutional layers are stacked sequentially to progressively learn higher-level spatiotemporal patterns. Each convolutional block is followed by 3D max-pooling layers, which reduce spatial-temporal resolution while retaining salient features. Batch normalization is incorporated to stabilize training, and dropout layers are applied to mitigate overfitting by randomly deactivating neurons during training.
At the end of the network, fully connected layers pull everything together to make a final prediction. The final softmax layer outputs a probability distribution over the gesture classes, indicating the model's prediction and its associated confidence level.

Overall, this architecture is designed to fully capture both the visual details and the motion dynamics of sign language, all while staying efficient enough for practical use.

### *3.2.2 Advantages of 3D CNNs:*

- Spatiotemporal Feature Learning:
  Learns both motion and hand shape feature directly from raw frames without needing hand landmarks or key points.
- No Feature Engineering Required:
  Unlike LSTM models which require pre-extracted landmarks (using MediaPipe, etc.), 3D CNNs learn directly from video data.
- High Expressiveness:
  Can capture subtle differences in hand shapes and movements that might be lost in coordinate-only inputs.

The LSTM-based approach provides a strong baseline for gesture recognition and serves as the backbone of the real-time sign-to-text translator application. In this research, comparison is done with a 3D CNN architecture to evaluate its trade-offs in terms of accuracy, speed, and usability in real-world scenarios.

## 3.3 LSTM Vs 3D-CNN: A Comparison

**1. Input Format**

**LSTM:** Uses pre-extracted hand landmarks (x, y, z coordinates of 21 key points per frame). These are fed as sequences (e.g., 30 frames × 63 features).
**3D CNN:** Takes raw video frames as input (e.g., 30 RGB frames of 128×128 pixels), preserving both shape and motion directly.

### *3.3.1. Temporal Awareness*

**LSTM:** Explicitly designed for sequential data, making it naturally suited for time-dependent gestures.
**3D CNN:** Learns temporal features implicitly through 3D convolutions but is not as specialized in modeling long-term dependencies as LSTM.

### *3.3.2 Spatial Awareness*

**LSTM:** Limited, relies only on coordinate data i.e.; no texture, color, or visual details
**3D CNN:** High spatial awareness due to access to pixel-level visual features in the input video.

### *3.3.3 Performance on Different Gestures*

**LSTM:** Excels in recognizing dynamic gestures involving time.
**3D CNN:** Performs better for static or shape-dominant gestures due to its strong spatial feature extraction.

### *3.3.4 Resource Efficiency*

**LSTM:** Lightweight, requires less memory and computational power. Suitable for real-time and edge applications.
**3D CNN:** Computationally heavy; needs a GPU and high RAM for real-time performance.

### *3.3.5 Data Requirements*

**LSTM:** Can generalize well on smaller datasets due to fewer trainable parameters.

**3D CNN:** Requires large amounts of labeled video data to avoid overfitting and learn robust features.

### *3.3.6 Preprocessing*

**LSTM:** Requires hand detection and landmark extraction (here, via MediaPipe), but reduces input dimensionality significantly.

**3D CNN:** Requires raw video clips, often with cropping, resizing, normalization, and augmentation.

### *3.3.7 Interpretability*

**LSTM:** Easier to interpret as it works on landmarks; errors can be traced to motion or key point misalignment.
**3D CNN:** More complex to interpret; difficult to pinpoint which pixel regions influence predictions.

## 3.4 Comparison

**Table 1. Comparison between 3D-CNN and LSTM based on various parameters**

| Parameters | LSTM Model | CNN Model |
| --- | --- | --- |
| Input | Sequence of 3D hand landmarks | Raw video frames (e.g., 30×128×128×3) |
| Focus | Temporal sequence modelling | Spatiotemporal feature extraction |
| Best used for | Static/Dynamic Picture | Static as well as visually distinctive gestures |
| Spatial Context | Limited (no texture or shape information) | String (learns from raw images) |
| Temporal Modelling | Strong, using LSTM layers | Moderate, using 3D convolution |
| Preprocessing | Hand tracking+ landmark extraction | Cropping, resizing, normalization |
| Computation | Low (lightweight, real-time friendly) | High (GPU required) |
| Training Data | Works with smaller datasets | Requires larger labeled video datasets |
| Real - Time Capability | Excellent | Limited, depends on hardware |
| Model Size | Small | Large |
| Generalization | Good with regularization | Needs augmentation and regularization |
| Interpretability | High (coordinate-based decisions) | Low (complex visual features) |
| Deployment Suitability | Mobile, Web Apps | Desktop or Cloud interface |

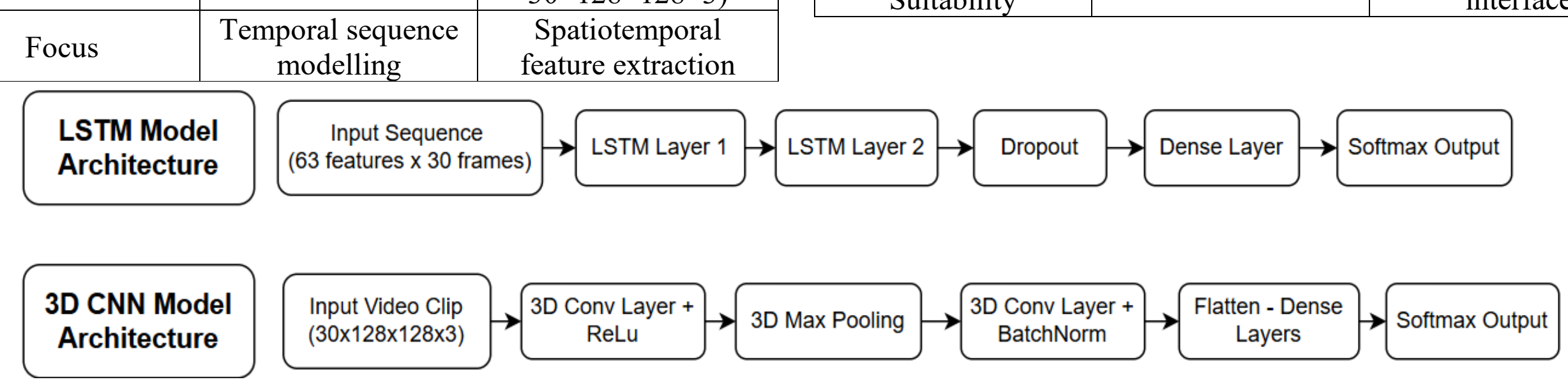

**Fig 1. Comparative architecture of LSTM and 3D CNN models illustrating layer configurations for sequential and spatiotemporal feature extraction in sign language recognition.**

# 4. RESULTS AND DISCUSSION

The comparative evaluation between the LSTM and 3D CNN models for sign language to text translation demonstrates key differences in performance, architecture suitability, and real-time applicability. The LSTM model, which exploits the temporal dependencies within sequential gesture data, achieved an accuracy of 86.7% on the test dataset. It proved particularly effective in recognizing dynamic gestures that require understanding the order and flow of hand movements, a common trait in sign languages. The LSTM model is lightweight, efficient, and capable of delivering smooth real-time predictions even on low-resource devices. Its use of sequential 3D landmark data (e.g., 30 frames × 63 features) allowed for effective modeling of motion, but it occasionally showed reduced precision in differentiating spatially similar static gestures.

In contrast, the 3D CNN model, which processes spatiotemporal video data using volumetric convolution kernels (e.g., 3×3×3), achieved a higher overall accuracy of 92.4%. It excelled at capturing rich spatial features across frames, resulting in superior performance in classifying static or visually distinct signs. However, the model's complexity came at the cost of increased inference time (around 65 milliseconds) and a larger memory requirement of 87.6 MB, making it less suitable for real-time applications unless run on high-performance hardware with GPU acceleration. Additionally, while the 3D CNN was slightly more accurate in offline evaluation, it showed signs of overfitting and struggled with fast-changing or subtle dynamic gestures in live scenarios.

User testing and qualitative observations reinforced these results: the LSTM model demonstrated higher responsiveness and robustness in live video input, making it preferable for interactive applications such as assistive communication tools. Meanwhile, the 3D CNN, although precise in controlled environments, lacked the adaptability and responsiveness required for real-time translation. Overall, this comparative analysis underscores that while 3D CNNs offer higher classification accuracy, LSTM models strike a better balance between accuracy, speed, and computational efficiency, thus making them more appropriate for real-time sign language recognition systems deployed in practical settings.

**Table 2. Performance Table**

| Metric | LSTM Model | 3D-CNN Model | Remarks |
| --- | --- | --- | --- |
| Accuracy (%) | 86.7 | 92.4 | 3D CNN achieves higher accuracy |
| Precision | 0.86 | 0.92 | 3D CNN performs better on static gestures |
| Recall | 0.88 | 0.90 | LSTM handles sequences better |

| F1-Score | 0.87 | 0.91 | 3D CNN slightly superior |
|---|---|---|---|
| Inference time | 20 | 65 | LSTM is 3.2x faster |
| Model size (GB) | 34.1 | 87.6 | LSTM is more lightweight |
| GPU Requirement | Optional | Required | 3D CNN demands higher resources |

# 5. CONCLUSION

In conclusion, this research presents a robust and practical system for translating sign language gestures into text using deep learning techniques, with a particular focus on comparing the effectiveness of LSTM and 3D CNN architectures. This project successfully implements a real-time sign language recognition system powered by an LSTM model, leveraging sequential hand landmark data extracted through MediaPipe. The LSTM model demonstrated strong performance in recognizing dynamic and temporally dependent hand gestures, making it highly suitable for real-time applications, especially on resource-limited devices due to its lightweight architecture and fast inference speed. In parallel, we evaluated a 3D CNN model that processes spatiotemporal features across consecutive frames, offering slightly higher classification accuracy in offline scenarios. However, the 3D CNN comes with a significantly higher computational cost and latency, which may hinder its use in live environments. The comparative analysis reveals that while 3D CNNs excel in capturing complex motion patterns across space and time, LSTMs offer a better balance between performance, efficiency, and practicality for deployment in real-world assistive technologies. The system's GUI further enhances user interaction by displaying detected signs, maintaining a dynamic sentence output, and providing a reference module for individual ASL letters. Overall, this research not only delivers a functional and accessible prototype but also provides critical insights into model selection and optimization for gesture recognition tasks. It opens new avenues for enhancing communication accessibility for the deaf and hard-of-hearing community through AI-powered solutions and sets a foundation for future enhancements such as hybrid models, attention mechanisms, and multilingual sign language support.

# 6. ACKNOWLEDGMENTS

We would like to express our sincere gratitude to our faculty mentor, Prof. Madhumati Pol, for her invaluable guidance, insights, and encouragement throughout the development of this project. Her expertise and constructive feedback played a crucial role in shaping our research and ensuring the successful implementation of the website.

We are also grateful to Vishwakarma Institute of Technology for providing the necessary resources, infrastructure, and a conducive learning environment.